\documentclass{article}

\usepackage{arxiv}

\usepackage[utf8]{inputenc} 
\usepackage[T1]{fontenc}    
\usepackage{hyperref}       
\usepackage{url}            
\usepackage{booktabs}       
\usepackage{amsfonts}       
\usepackage{nicefrac}       
\usepackage{microtype}      
\usepackage{lipsum}		
\usepackage{graphicx}
\usepackage{natbib}
\usepackage{doi}
\usepackage{subfigure}
\usepackage{graphicx}
\usepackage{amsmath,amssymb} 
\usepackage[ruled,linesnumbered]{algorithm2e}
\usepackage{epstopdf}
\usepackage{color}

\title{Robust Correlation Tracking via Multi-channel Fused Features and Reliable Response Map}

\date{7 May, 2018}	

\author{ {\hspace{1mm}Xizhe Xue}\\
	School of Computer Science\\
	Northwestern Polytechnical University\\
	Xi'an, China \\
	\texttt{xuexizhe@mail.nwpu.edu.cn} \\
	\And
	{\hspace{1mm}Ying Li}\thanks{Corresponding author.}  \\
	School of Computer Science\\
	Northwestern Polytechnical University\\
	Xi'an, China \\
	\texttt{lybyp@nwpu.edu.cn} \\
	\And
	{\hspace{1mm}Qiang Shen} \\
	Department of Computer Science\\
	Aberystwyth University\\
	Aberystwyth, SY23 3DB, UK \\
	\texttt{qqs@aber.ac.uk} \\
}

\renewcommand{\shorttitle}{Robust Correlation Tracking via Multi-channel Fused Features and Reliable Response Map}

\hypersetup{
pdftitle={Robust Correlation Tracking via Multi-channel Fused Features and Reliable Response Map},
pdfsubject={Computer vision},
pdfauthor={Xizhe Xue, Ying Li},
pdfkeywords={Visual tracking, Correlation filter, Feature fusion},
}

\begin{document}
\maketitle

\begin{abstract}
	Benefiting from its ability to efficiently learn how an object is changing, correlation filters have recently demonstrated excellent performance for rapidly tracking objects. Designing effective features and handling model drifts are two important aspects for online visual tracking. This paper tackles these challenges by proposing a robust correlation tracking algorithm (RCT) based on two ideas: First, we propose a method to fuse features in order to more naturally describe the gradient and color information of the tracked object, and introduce the fused features into a background aware correlation filter to obtain the response map. Second, we present a novel strategy to significantly reduce noise in the response map and therefore ease the problem of model drift. Systematic comparative evaluations performed over multiple tracking benchmarks demonstrate the efficacy of the proposed approach.
\end{abstract}

\keywords{Visual tracking \and  Correlation filter \and Feature fusion}
\section{Introduction}
\label{sec:intro}
Visual tracking plays an active role in a wide range of applications, including driverless vehicles, robotics, surveillance, human-computer interaction, surveillance, motion analysis. Recent years have witnessed significant developments in visual tracking, where an enormous amount of research effort has gone into tasks such as short-term single-object tracking where the target appears in the first frame. However, many challenges remain, such as deformation, target rotation, scale variation and fast motion.

Correlation filters (CFs) have recently been introduced for visual-tracking and have been shown to achieve high speed as well as robust performance~\cite{bertinetto2016staple,bolme2010visual,henriques2012exploiting,boddeti2013correlation,danelljan2017discriminative,danelljan2015learning,galoogahi2017learning,danelljan2017eco}. For instance, David S. Bolme et al.~\cite{bolme2010visual} proposed one based on the minimum output sum of squared errors (MOSSE), which works by manipulating the maximum cross-correlation response between the model and the candidate patch. Henriques et al.~\cite{henriques2012exploiting} exploited the circulate structure and Fourier transformation in a kernel space (so-called the CSK tracker), offering excellent performance on a range of computer vision problems. A vector correlation filter (VCF) was devised by Boddeti et al.~\cite{boddeti2013correlation} to minimize localization errors while improving the tracking speed. The DSST tracker~\cite{danelljan2017discriminative} employs adaptive multi-scale correlations filters using HOG features in an effort to handle scale changes in target objects. To learn a model that is inherently robust to both color changes and deformations, Staple~\cite{bertinetto2016staple} combines two image patch representations that are sensitive to potential competing factors such as rotation and deformation. Danelljan et al.~\cite{danelljan2015learning} utilized a spatial-regularization component to penalize correlation filter coefficients as a function of their spatial location. The work in~\cite{galoogahi2017learning} presents a background-aware correlation filter (BACF) that can model how the background, as well as the foreground, of an object may vary over time. To drastically reduce the number of modeling parameters, ECO~\cite{danelljan2017eco} introduces a factorized convolution operator into traditional CF, where a compact generative model on the training sample distribution significantly reduces the memory and time complexity, while enhancing sample diversity.

Recognizing the success of deep convolutional neural networks (CNNs) on a wide range of visual-recognition tasks, a number of tracking methods based on deep features and correlation filters have been developed ~\cite{ma2015hierarchical,qi2016hedged}. Empirical studies using large object-tracking benchmark datasets show that such CNN-based trackers perform favorably against methods based on the use of hand-crafted features. However, extracting CNN features from each frame, and training or updating CF trackers over high dimensional deep features, is computationally expensive. Therefore for visual tracking, such an approach often leads to poor real-time performance.

\textbf{Contribution}: In this paper, we propose a robust correlation tracking method (RCT) via the exploitation of feature fusion and reliable response. A fused feature herein describes the gradient and color information conjunctively in a more natural way as compared to existing techniques. The novel fused features are then embedded into a correlation filter that is background-aware (in the sense that the filter is capable of learning from real, negative examples densely extracted from the background). In addition, an adaptive optimization strategy is introduced to remove the untrusted part of the response map that is caused by deformation or other challenging factors.

We evaluate the proposed tracker on the OTB50, OTB100 and Temple-Color128 datasets. The results demonstrate that our method obtains a very competitive accuracy level in comparison with the state-of-the art CF-based trackers, but does so with a real-time tracking speed of 17 FPS on a standard desk-top CPU.
\section{Related Work}
This section first provides an overview of feature representation in correlation tracking and then outlines the existing methods for addressing model drift that are most relevant to the present work.

\noindent\textbf{Feature representation in correlation tracking.} It is critical to employ an efficient feature-representation mechanism in order to improve performance during object-tracking. Gradient and color features are the most popular single types of feature. In particular, color features help capture rich color characteristics, while features reflecting the histogram of oriented gradient (HOG)~\cite{dalal2005histograms} are adept in capturing abundant gradient information. Based on these feature descriptions, a variety of techniques on target-tracking have been proposed. For instance, CSK~\cite{henriques2012exploiting} exploits the circulant structure while only utilizing the grey-level features over single channels. Upon this basis, the kernelized correlation filter (KCF)~\cite{henriques2015high} employs HOG features to improve filter performance. Danelljan et al.~\cite{danelljan2014adaptive} investigated color attributes to reinforce correlation-filter tracking, drawing the conclusion that using CN features can greatly improve the performance of correlation filters. Also, the powerful features including HOG and color-naming have been integrated together to further boost the overall tracking performance in~\cite{li2014scale}. Furthermore, applications of features extracted by CNNs~\cite{ma2015hierarchical} have been widely studied, preliminarily in performing the task of visual tracking, but are still limited due to expensive computational costs incurred. As such, whilst the aforementioned techniques have shown good potential, how to jointly utilize different features to increase the tracking performance remains an open issue.

\noindent\textbf{Robustness to model drift.} Model drifts lead to inaccurate model-based predictions. In addressing this problem, Kalal et al.~\cite{kalal2012tracking} proposed an approach that decomposed the ultimate task of tracking into subtasks of tracking, learning and detection (TLD), where tracking and detection reinforce each other. Also, the technique of Multiple-Instance Learning~\cite{babenko2011robust}(MILTrack) was introduced to train with bags of positive examples. Recently, discriminative correlation filters (DCF) which relies on a periodic assumption of the training and detection samples have been shown to be able to achieve tracking by employing a circular correlation. However, the circulant shifted samples in such trackers suffer from periodic repetitions on boundary positions, thereby leading to model drift and significantly degrading the tracking performance. Spatial regularization methods have since been suggested to alleviate the unwanted boundary effects. For example, using the Alternating Direction Method of Multipliers (ADMM), Galoogahi et al.~\cite{galoogahi2015correlation} resolved a constrained optimization problem for single-channel DCF. Somewhat differently, the SRDCF formulation~\cite{danelljan2015learning} allows correlation filters to be trained on a significantly larger set of negative training samples, without corrupting the positive samples, where a spatial regularization component is introduced to the training process to penalize the correlation filter coefficients in relation to their spatial location. Unlike previous DCF-based trackers in which negative examples are restricted to circular shifted patches, BACF~\cite{galoogahi2017learning} utilizes a correlation filter whose spatial size is much smaller than that of the training samples; real negative training examples, densely extracted from the background are utilized. These works have aimed to prevent model drift through modifying the training strategy rather than improving the underlying model-based predictions themselves. In the present work, we improve the predictions by obtaining and manipulating a more reliable response map, leading to an enhanced tracking result.

Compared with the existing methods, our proposed tracker has several merits: (1) while RCT may be viewed as an (improved) approximation to the work of~\cite{galoogahi2017learning} on multiple training samples, the filter works more efficiently owing to the use of a more reliable response map; and (2) with the introduction of fused features, the RCT tracker can learn more robust features than the previous work, thereby leading to superior tracking performance.
\begin{algorithm}
	\caption{Tracking algorithm}\label{algorithm}
	\KwData{Frames $I_f$, initial target location $p_1$ ( $f$ means the number of the current frame)}
	\KwResult{Target location $p_f$}
	\textbf{repeat}\;
	Crop an image region from $I_f$ at the last location and extract its fused-feature vector $x_f$\;
	Compute the optimum correlation filter (via Eq.2) and obtain the original response map\;
	Construct the mask to yield a reliable response map\;
	Detect the target location $p_f$ via the reliable response map\;
	Estimate the scale of the target and update the tracking model (as summarized in Section 3.4)\;
	\textbf{until} end of video sequence\;
\end{algorithm}
\section{Proposed Approach}
We aim to develop a robust tracking algorithm that is adaptive to significant changes without being prone to drifting. We first propose a fused feature mechanism which describes the gradient and color information in an integrated way. Then, a background-aware correlation filter based on the exploitation of fused features is designed to obtain a response map. Furthermore, the mask obtained according to the value of the response map will be multiplied with a given original image to form a more reliable response map, which help alleviate possible model drifts. The proposed method is presented in Algorithm \ref{algorithm}.

\subsection{Multi-channel Fused Features}
Features play an important role in computer vision; much of the impressive progress in object detection can be attributed to the improvement in the representation power of features. Gradient and color features are the most widely exploited in object detection and tracking. Indeed, previous work~\cite{khan2013discriminative} has verified that there exists a strong complementarity between gradient and color features.

How to best utilize different features jointly for visual tracking is still an open question. In the medical profession, Duplicity theory~\cite{hecht1935intermittent} concerns the comparison (in terms of both differences and similarities) and interaction between the cone and rod systems in the visual pathways, with the assumption that the cone system is active during daylight vision and the rod system functions in low light (night time). The correct (whether cone or rod) system is first selected according to the luminance of environment; then, the chosen system will work to obtain detailed information of the surroundings.

\begin{figure}
	\centering
	\includegraphics[width=0.8\linewidth]{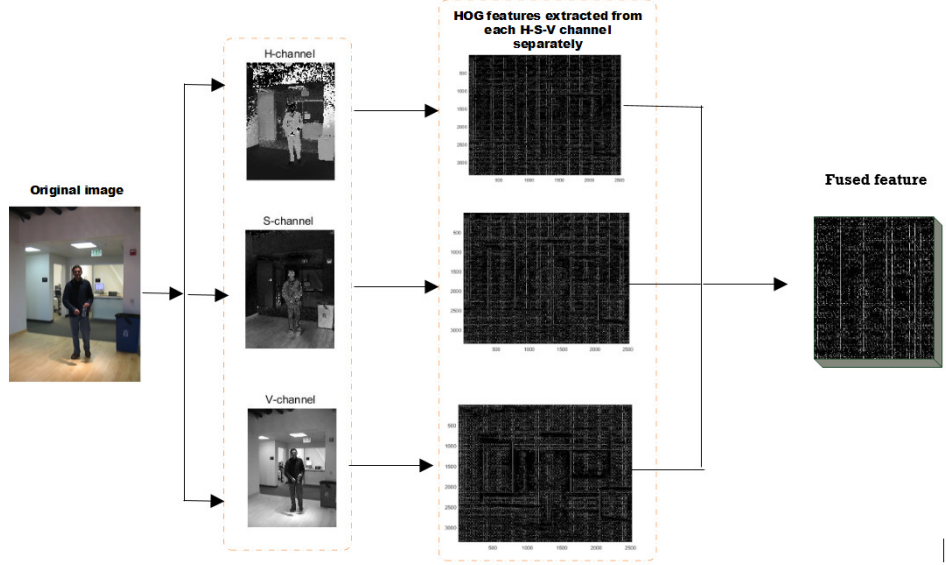}
	\caption{Extraction of our fused features.}
	\label{fig:feature}
\end{figure}

Inspired by the duplicity theory of vision, we construct a more natural feature representation. In our setting, instead of concatenating the color and gradient features directly, we first transform the original image into HSV (Hue, Saturation, Value) color space, which is based more upon how colors are organized and conceptualized in human vision. In such a color space, brightness and colorfulness are absolute measures, which usually describe the spectral distribution of light entering the eye. Benefiting from this, our fused feature performs robust to the illumination variation. Secondly, HOG gradient information is extracted from each channel of the HSV color space, separately. Finally, all the HOG features are concatenated to form our proposed fused feature, in the form of a matrix. Hence, for terminology, we think of the output feature as the combination of fused-(input)-features or, as a (singular) fused feature. Without losing generality and for conciseness, we term the resultant feature descriptor a fused feature. The process  to extract the final fused feature is shown in Figure \ref{fig:feature}.
\subsection{CF Tracking through Fused Features}
In this section, we introduce our fused feature into background-aware correlation filter \cite{galoogahi2017learning} to construct a better correlation-tracking framework. We utilize a correlation filter with a spatial size which is smaller than the size of training examples to reduce the boundary effects. Denote $x_k$ as the fused feature vector of a cardinality ${x_k}\in {R^T}$ , respectively. We consider $y \in {R^T}$ as the desired correlation output corresponding to a given sample $x_k$  . A correlation filter $w$ with the dimensionality of $D$ (where $T > >D$) is then learned by solving the following minimization problem:
\begin{align}
E(w) = \sum\limits_{j = 1}^T {||{y_j} - \sum\limits_{k = 1}^K {{w_k}^{\top}\cdot{\rm{P}}{x_k}[\Delta {\tau _j}]} |{|^2} + \lambda \sum\limits_{k = 1}^K {||{w_k}||_2^2}}
\label{Eq:1}
\end{align}
where $\lambda$ is a regularization parameter, and ${\rm{P}}{x_k}[\Delta {\tau _j}]$ generates all circular shifts of size $D$ from the entire frame over all $j = \left[ {0,...,T - 1} \right]$ steps.

Note that the Eq.\ref{Eq:1} can be readily transformed into frequency domain  (using discrete Fourier transform) in order to improve the computational efficiency. We introduce $\hat g = {[\hat g_1^{\rm T}, \cdot  \cdot  \cdot ,\hat g_K^{\rm T}]^{\rm T}}$ as an auxiliary variable. The trained filter in the frequency domain will be written as
\begin{align}
\begin{array}{l}
E(w,\hat g) = ||\hat y - \hat X\hat g||_2^2 + \lambda ||w||_2^2\\
s.t.\hat g = \sqrt T ({\rm{F}}{{\rm{P}}^{\top}} \otimes {{\rm{I}}_k})w
\end{array}
\label{Eq:2}
\end{align}
where $\hat X$ is denoted by $\hat X = {[diag{({\hat x_1})^{\top}},...,diag{({\hat x_k})^{\top}}]^{\top}}$, $I_k$ is the $K\times K$ identity matrix, and $\otimes$ denotes the Kronecker product. In particular, $\hat A $ represents the Discrete Fourier Transform (DFT) of a signal $A$,where $F$ is the orthonormal $T\times T$ matrix of complex basis vectors, mapping any $T$-dimensional vectorized signal to its Fourier domain. The transpose operator $\top$ on a complex vector or matrix gives the conjugate transpose.

By directly employing the Augmented Lagrangian Method (ALM)~\cite{galoogahi2017learning}, we can solve Eq.\ref{Eq:2} and obtain the required correlation filter ${\hat g^{(f - 1)}}$, where $f$ is the current frame number.
\subsection{Object Location by Reliable Response}
Representing the response value of every pixel, the response map $R$ in frame $f$ can be computed by applying the filter ${\hat g^{(f - 1)}}$ that has been updated in the previous frame. Due to the challenges typically faced in performing real-world tracking tasks, such as deformation and rotation, the similarity between the target and modeling template may be decreased, leading to great risk of model drift or locating mistakenly.
\begin{figure}
	\centering
	\includegraphics[width=0.8\linewidth]{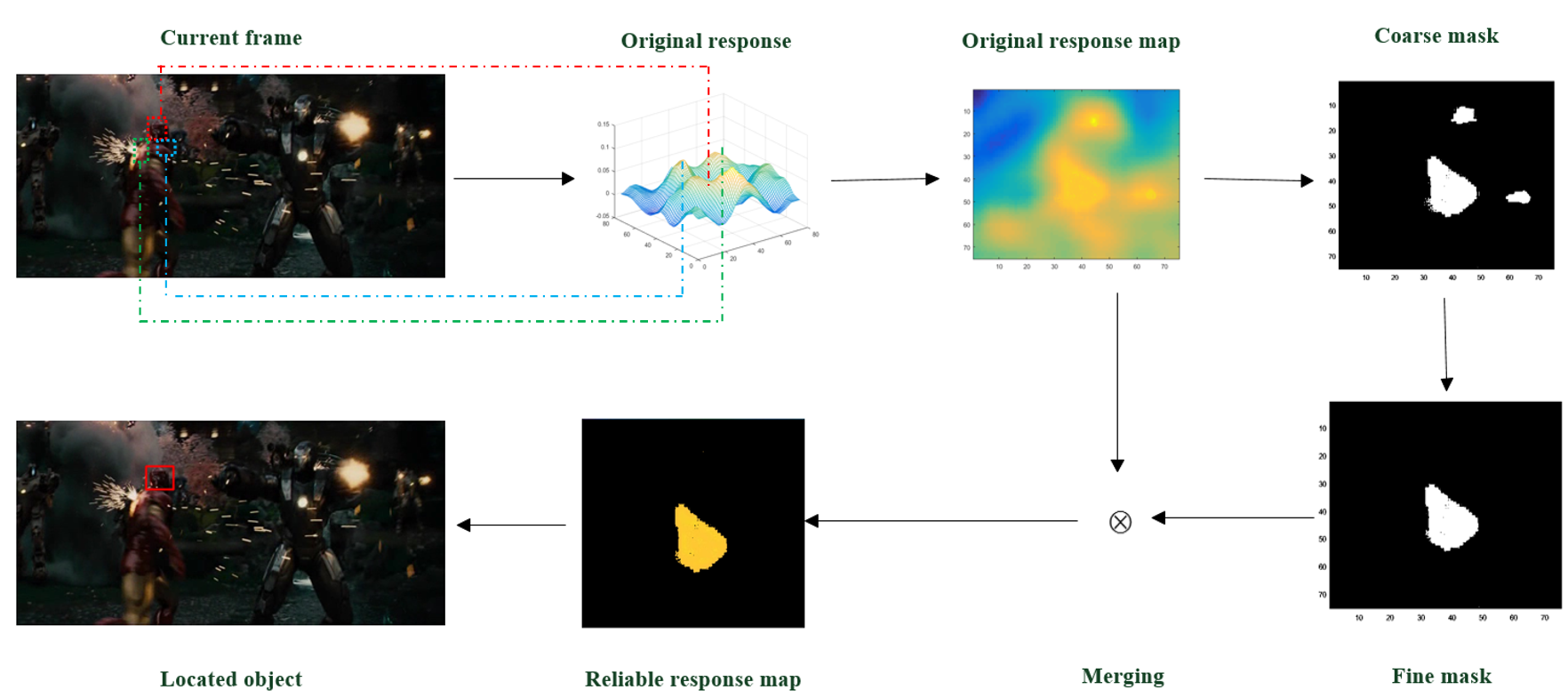}
	\caption{Object location by reliable response.}
	\label{fig:reliable_response_map}
\end{figure}

How to remove a lot of potentially misleading redundant information (responses to similar objects) contained in the original response map then? As shown in Figure \ref{fig:reliable_response_map}, when noise exists, the position with the maximum value in the response map does not necessarily correspond to the real target. In this case, simply taking the position with the highest response as the target position is rather unreliable. Through a large number of experiments, empirically we find that the response peak of the real target often changes gradually, while the response peak of the disturbed object is usually very steep and looks very abrupt. Accordingly, in order to exclude the anomalies, we first try to identify the target proposals which are associated with a relatively high value in the response map. In order to achieve this goal, we set a threshold $Th$, which divides the response map into two parts. The pixels with a gray value greater than $Th$ belong to the target proposal set $A$ and the remaining are deemed to attribute to the background part $B$. The number of pixels contained in the two parts is represented with $NA$ and $NB$ respectively. We vary $Th$ from 0 to 255, each time, $NA$ and $NB$ are counted to calculate the ratio of the target proposals $P_f$ in the current frame, such that
\begin{align}
\begin{array}{l}
{P_f} = \frac{{{N_A}}}{{{N_A} + {N_B}}}
\end{array}
\label{Eq:3}
\end{align}
Repeat Eq.\ref{Eq:3} until the difference between the $P_f$ and a predefined value $P$ is less than the error range $G$,
\begin{align}
\begin{array}{l}
|{P_f} - P| < G
\end{array}
\label{Eq:4}
\end{align}
When Eq.\ref{Eq:4} is satisfied, the grey value of pixels in the set $A$ is reset to 255, while each of the rest pixels is set to 0. From this, a number of connected domains are obtained. Then, any connected domain whose area is less than a fixed threshold $AR$ is deleted to form the fine mask $K$. By multiplying $K$ with the original response map $R$, the reliable response map results. Finally the position with the maximum value in the reliable response map is treated and recognized as the target location.
\subsection{Model Updating and Scale Estimation}
To obtain a robust approximation, we update the model of the correlation filter $M$ at the $f-th$ frame using a moving average:
\begin{align}
\begin{array}{l}
M_{}^f = (1 - \eta ){M^{f - 1}} + \eta {M^f}
\end{array}
\label{Eq:5}
\end{align}
where $\eta$ is the learning rate.

In order to be adaptable to any change of the scale of a target, the filter is applied on multiple resolutions of the searching area where tracking takes place~\cite{danelljan2015learning}. The searching area has the same spatial size as that of the filter ${\hat g}$ . This returns $S$ correlation outputs with different scales, where $S$ is the number of scales. The scale with the maximum correlation output is used to update the object location and the subsequent scale.
\section{Experimental Results}
In order to present an objective evaluation regarding the performance of the proposed approach, we examine our RCT tracker on three standard datasets,
including OTB50~\cite{WuLimYang13}, OTB100~\cite{wu2015object}, and Temple-Color128 (TC128)~\cite{Liang2015Encoding}. Both the general capability and the special scenarios-handling ability are tested. We compare our tracker with a range of excellent state-of-the-art trackers, including: SRDCFdecon \cite{danelljan2016adaptive}, MCPF \cite{zhang2017multi}, ECO\_HC \cite{danelljan2017eco}, BACF \cite{galoogahi2017learning}, CF\_AT \cite{bibi2016target}, CACF \cite{mueller2017context}, BIT \cite{cai2016bit}, fDSST \cite{danelljan2017discriminative}, Staple \cite{bertinetto2016staple}. Different metrics may be used for evaluation depending on preferred perspectives, amongst which one-pass evaluation (OPE) is arguably the most commonly used evaluation method. OPE runs a tracker on each sequence once: it initializes a tracker using the ground truth object state in the first frame, and reports the average precision or success rate of all subsequent results. Having recognized this, OPE is also used herein to comparatively evaluate the present work.
\subsection{Implementation Details}
We evaluate our tracker on the challenging Visual Tracker Benchmark \cite{wu2015object}, by following rigorously the existing evaluation protocols. The experiments are performed in Matlab2014a on an Inteli5 3.2GHz CPU with 4G RAM. In all the experiments carried out, we use the same parameter values for all image sequences. We employ 31-channel HOG features using 4 x4 cells to obtain the fused feature. The regularization factor is empirically set to 0.001 and the number of scales is set to 5 with a scale-step of 1.01. A 2D Gaussian function with bandwidth of $\sqrt {wh/16}$ is used to define the correlation output for an object of size $[h,w]$. The learning rate ($\eta$ ) of the correlation filter is 0.013.The area threshold $AR$ is set to 20\%, and the error range $G$ is set to be within 10.
\subsection{Evaluation on OTB Dataset}
We implement the one-pass experiment on the OTB50 and OTB100 benchmark datasets. All these image sequences are annotated with 11 attributes which cover various challenging factors, including scale variation (SV), occlusion (OCC), illumination variation (IV), motion blur (MB), deformation (DEF), fast motion (FM), out-of plane rotation (OPR), background clutters (BC), out-of-view (OV), in-plane rotation (IPR) and low resolution (LR).

Let $r_T$ represent the tracking result of trackers and  $r_G$ denote the given groudtruth, the success plots below show the ratios of successful frames whose overlap score ($OS$) is defined as $OS{\rm{ =  }}\frac{{{r_T} \cap {r_G}}}{{{r_T} \cup {r_G}}}$. The trackers are ranked by precision values (with a threshold of 20) in the precision plots, and by the area-under-curve (AUC) depicted by the success plots. If $OS>0.5$, the tracking result is deemed correct.
\begin{figure}
	\centering
	\subfigure[\tiny{Success plots on OTB50}]{
		\label{fig:subfig:a} 
		\includegraphics[width=0.235\linewidth]{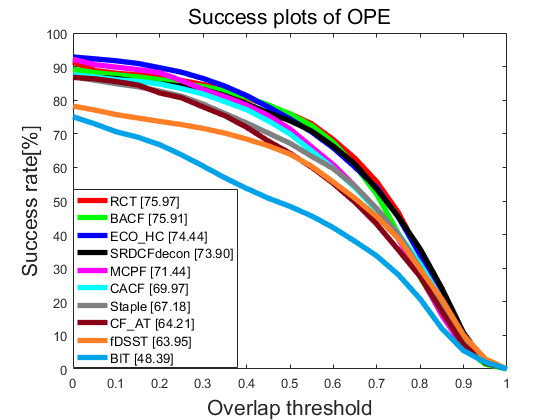}}
	\subfigure[\tiny{Precision plots on OTB50}]{
		\label{fig:subfig:b} 
		\includegraphics[width=0.235\linewidth]{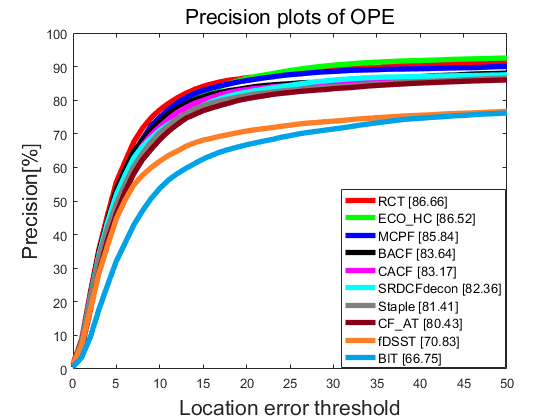}}
	\subfigure[\tiny{Success plots on OTB100}]{
		\label{fig:subfig:a} 
		\includegraphics[width=0.235\linewidth]{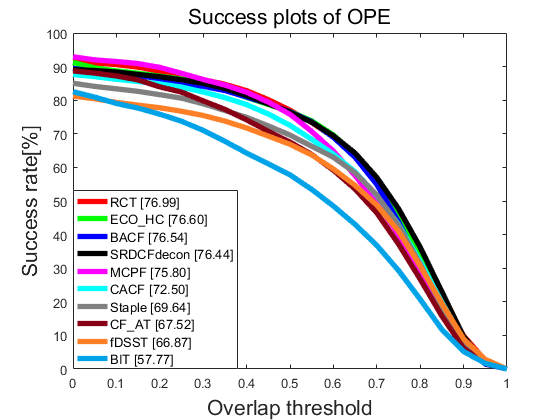}}
	\subfigure[\tiny{Precision plots on OTB100}]{
		\label{fig:subfig:b} 
		\includegraphics[width=0.235\linewidth]{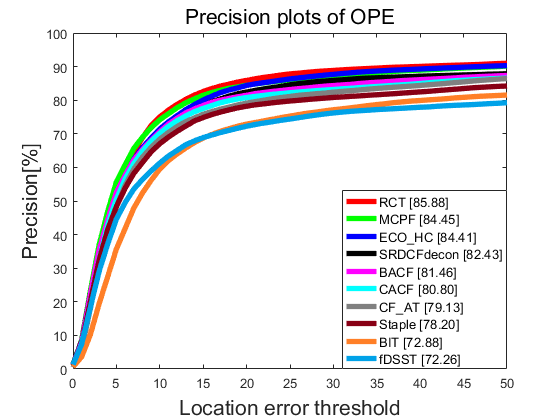}}
	\caption{Results of the proposed tracker and other compared trackers on OTB dataset.}
	\label{fig:OTB}
\end{figure}

Figures \ref{fig:OTB} shows the success and precision plots on the OTB50 and OTB100 dataset, respectively. Only the top 10 trackers are displayed for clarity. Overall, our RCT is better than the other state-of-the-art trackers and achieves a significant gain in the OTB100 dataset. BIT is a tracker that extracts low-level biologically-inspired features while imitating an advanced learning mechanism to combine generative and discriminative models for target location. Our RCT improves on BIT by an average of 27.58\% in the AUC scores on OTB50, and by 19.22\% on OTB100. This testifies to the extraordinary performance of the fused features. Since ECO\_HC ranks first amongst all of the HOG-based trackers, it makes for a good representation of the existing trackers compared. Our RCT improves on ECO\_HC in the average AUC scores on both datasets. The BACF and SRDCFdecon trackers primarily focus on the boundary effects, with reported speeds of 30 FPS and 1 FPS respectively. Our RCT employs the ''background-aware'' mechanism from BACF, but performs favorably over BACF as well as SRDCFdecon.
\subsection{Evaluation on TC128 dataset}
\begin{figure}
	\centering
	\subfigure[Success plots]{
		\label{fig:subfig:a} 
		\includegraphics[width=0.44\linewidth]{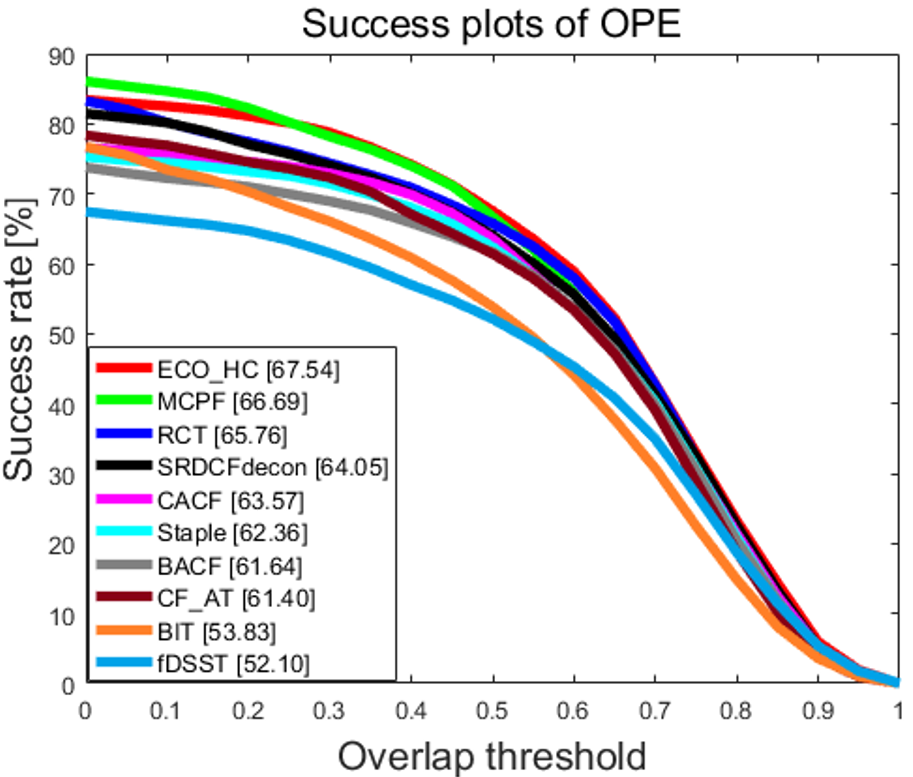}}
	\subfigure[Precision plots]{
		\label{fig:subfig:b} 
		\includegraphics[width=0.45\linewidth]{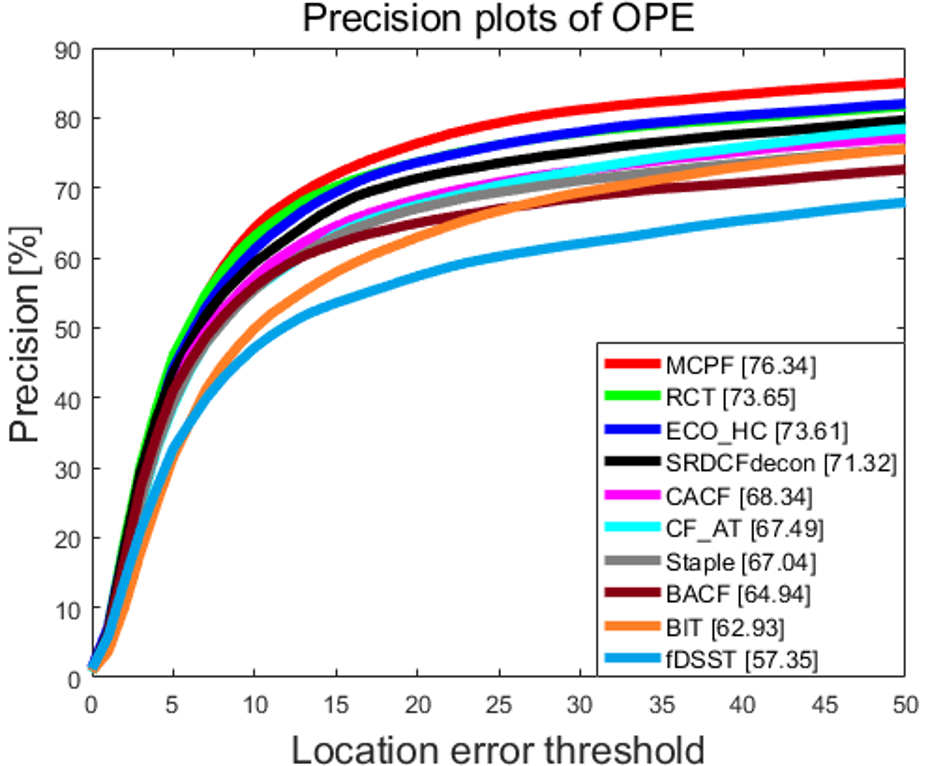}}
	\caption{Results of proposed tracker and other compared trackers on TC128 dataset.}
	\label{fig:TC128}
\end{figure}

TC128 is the first comprehensive color-tracking benchmark. The results of ten trackers on the 128 videos are summarized in Figure \ref{fig:TC128}, which shows the average rates of both the success plots and the precision plots. As can be seen from these results, our tracker always performs reliably and can achieve the optimal or at the least, very close to the best with respect to both metrics. Note that MCPF (which ranks highly) utilizes deep features obtained using the VGG-19 convolutional neural network. Extracting CNN-features from each frame and training or updating CF trackers over high dimensional deep features is computationally expensive (which results in poor real-time performance). Although the average AUC scores of our tracker is a little weaker compared with MCPF, it runs more than 34 times faster as compared to MCPF, which has a reported speed of 0.5 FPS.
\subsection{Further State-of-the-art Comparison}

Table \ref{tab:1} compares our method with the state-of-the-art CF based trackers on the OTB50, OTB100 and TC128 datasets, where our RCT achieves the highest accuracy across all HOG-based trackers and, including even MPCF, which utilizes deep features (and hence involves substantially more computation). More specifically, as shown in this table, RCT achieves the best AUC (76.99\%) on OTB100 followed by ECO\_HC (76.60\%) and SRDCFdecon (76.44\%). On the OTB50 dataset, RCT (75.96\%) outperforms ECO\_HC (74.44\%) and SRDCFdecon (73.90\%). Finally, our RCT (72.92\%) is ranked first on average, closely followed by ECO\_HC (72.86\%), which is the winner of the comparison on TC128.
\begin{table}
	\large
	\centering
	\caption{Success rates (OS$>$0.50) of the proposed trackers versus other state-of-art trackers. The first, second and third best methods are shown in color (The red ranks first, green means second and blue means third).}
	\begin{tabular}{ccccc}
		\hline
		\multicolumn{1}{r}{} & \multicolumn{1}{p{4em}}{OTB50} & \multicolumn{1}{p{4em}}{OTB100} & \multicolumn{1}{p{3em}}{TC128} & \multicolumn{1}{p{6em}}{Avg.succ.rate}  \\
		\hline
		SRDCFdecon & \textcolor[rgb]{ 0,  .69,  .941}{73.90} & \textcolor[rgb]{ 0,  .69,  .941}{76.44} & 64.05  & \textcolor[rgb]{ 0,  .69,  .941}{71.46}  \\
		MCPF  & 71.44 & 75.80  & \textcolor[rgb]{ 0,  .69,  .314}{66.69} & 71.31  \\
		ECO\_HC & \textcolor[rgb]{ 0,  .69,  .314}{74.44}  & \textcolor[rgb]{ 0,  .69,  .314}{76.60} & \textcolor[rgb]{ 1,  0,  0}{67.54} & \textcolor[rgb]{ 0,  .69,  .314}{72.86}  \\
		BACF  & 75.91  & 76.54  & 61.64  & 73.36  \\
		CF\_AT & 64.21  & 67.52  & 61.40 & 64.37   \\
		CACF  & 69.97  & 72.50  & 63.57  & 68.68   \\
		BIT   & 48.39  & 57.77  & 53.83  & 53.33  \\
		fDSST & 63.95  & 66.87  & 52.10  & 60.97   \\
		Staple & 67.18  & 69.64  & 62.36  & 68.39  \\
		RCT    &  \textcolor[rgb]{ 1,  0,  0}{75.97} & \textcolor[rgb]{ 1,  0,  0}{76.99} & \textcolor[rgb]{ 0,  .69,  .941}{65.76} & \textcolor[rgb]{ 1,  0,  0}{72.91}  \\
		\hline
	\end{tabular}%
	\label{tab:1}%
\end{table}%
\section{Conclusion}
In this paper, we have proposed a robust correlation filter-based tracking method via the use of multi-channel fused features and reliable response maps. The correlation filter that utilizes multi-channel fused features leads to a significant improvement in tracking performance while dealing with challenging factors such as deformation and scale variation. We have also proposed a novel strategy to obtain a more reliable response map, thereby locating the target through it. This allows our tracker to reduce the probability of incorrect locating when target occlusion and rotation exist severely. Comparative experimental investigations have proven, both quantitatively and qualitatively that our approach outperforms the state-of-the-art tracking methods. Future work will involve investigating more powerful fused features that combine intensity and color information while keeping the speed sufficiently fast for real-time applications.

\bibliographystyle{unsrtnat}
\bibliography{references}  

\end{document}